\documentclass[a4paper,fleqn]{cas-dc}
\makeatletter
\def\ps@first{%
   \let\@oddhead\@empty
   \let\@evenhead\@empty
   \def\@oddfoot{}
   \let\@evenfoot\@oddfoot
}
\usepackage{fancyhdr}
\usepackage{etoolbox}

\usepackage{setspace}      
\usepackage[switch]{lineno}

\usepackage{natbib}

\def\tsc#1{\csdef{#1}{\textsc{\lowercase{#1}}\xspace}}
\tsc{WGM}
\tsc{QE}
\tsc{EP}
\tsc{PMS}
\tsc{BEC}
\tsc{DE}

\usepackage{etoolbox}

\begin{document}
\let\WriteBookmarks\relax
\def\floatpagepagefraction{1}
\def\textpagefraction{.001}

\title [mode = title]{Development and Deployment of a Big Data Pipeline for Field-based High-throughput Cotton Phenotyping Data}                      

\tnotetext[1]{This work was supported by the United States National Institute of Food and Agriculture (NIFA) under award no. 2020-67021-32461.}

%
\author[1]{Amanda Issac}[type=editor]

\cormark[1]

\address[1]{School of Electrical and Computer Engineering, University of Georgia, 
    Athens,
    Georgia,
    United States of America}

\author[2]{Alireza Ebrahimi}
\cormark[1]
\address[2]{Department of Mechanical Engineering, Clemson University,
    Clemson,
    South Carolina,
    United States of America}

\author[2]{Javad {Mohammadpour Velni}}
\cormark[1]
    
\author[3]{Glen Rains}
\cormark[1]
\address[3]{Department of Entomology, University of Georgia, 
    Tifton,
    Georgia,
    United States of America}


\begin{abstract}
In this study, we propose a big data pipeline for cotton bloom detection using a Lambda architecture, which enables real-time and batch processing of data. Our proposed approach leverages Azure resources such as Data Factory, Event Grids, Rest APIs, and Databricks. This work is the first to develop and demonstrate the implementation of such a pipeline for plant phenotyping through Azure's cloud computing service. The proposed pipeline consists of data preprocessing, object detection using a YOLOv5 neural network model trained through Azure AutoML, and visualization of object detection bounding boxes on output images. The trained model achieves a mean Average Precision (mAP) score of 0.96, demonstrating its high performance for cotton bloom classification. We evaluate our Lambda architecture pipeline using 9,000 images yielding an optimized runtime of 34 minutes. The results illustrate the scalability of the proposed pipeline as a solution for deep learning object detection, with the potential for further expansion through additional Azure processing cores. This work advances the scientific research field by providing a new method for cotton bloom detection on a large dataset and demonstrates the potential of utilizing cloud computing resources, specifically Azure, for efficient and accurate big data processing in precision agriculture. 
\end{abstract}



\maketitle
\section{Introduction}
The demand for sustainable agriculture has put significant pressure on the agriculture sector due to the rapid growth of the global population. Precision farming techniques enabled by Computer Vision (CV) and Machine Learning (ML) have emerged as promising solutions where crop health, soil properties, and yield can be monitored and lead to efficient decision-making for agriculture sustainability. Data would be gathered through heterogeneous sensors and devices across the field like moisture sensors and cameras on the rovers. However, the huge number of objects in farms connected to the Internet leads to the production of an immense volume of unstructured and structured data that must be stored, processed, and made available in a continuous and easy-to-analyze manner \citep{Gilbertson2017}. Such acquired data possesses the characteristics of high volume, value, variety, velocity, and veracity, which are all characteristics of big data. In order to leverage the data for informed decisions, a big data pipeline would be needed.

One area of agriculture that faces particular challenges with regard to yield prediction is cotton production. The operation of cotton production is faced with numerous challenges, a major one being the timely harvesting of high-quality cotton fiber. Delayed harvesting can lead to the degradation of cotton fiber quality due to the exposure to unfavorable environmental conditions.  Therefore, to avoid degradation, it is vital for harvesting cotton when at least 60\% to 75\% are fully opened, but also prior to the 50-day benchmark when bolls begin to degrade in quality.  \citep{uga2019georgia}. In addition, cotton harvesting is costly, as the machines used for their processing can weigh over 33 tons and can also cause soil compaction, hence reducing land productivity \citep{antille2016soil}. Finally, a lack of skilled labor and external factors such as climate change, decreasing arable land, and shirking water resources hinder sustainable agricultural production \citep{food2009global}. In this context, heterogeneous and large-volume data is collected using various static and moving sensors. Therefore, it is imperative to develop a platform that can handle real-time streams and manage large datasets for High-Throughput Phenotyping (HTP) applications. However, most conventional storage frameworks adopted in previous studies support only batch query processing and on-premise servers for data processing. Rather than implementing on-premise processing, the adoption of cloud computing can help prevent over- or under-provisioning of computing resources, reducing costly waste in infrastructure for farmers as shown in \citep{Kiran2015} who introduced a cost-optimized architecture for data processing through AWS cloud computing resources. Therefore, leveraging cloud computing could be a viable option for developing an efficient and scalable platform for HTP applications.

In this paper, we aim to implement batch and real-time processing using cloud computing which can help prevent over- or under-provisioning of computing resources. For that, we propose a big data pipeline with a Lambda architecture through Azure which allows for the cohesive existence of both batch and real-time data processing at a large scale. This two-layer architecture allows for flexible scaling, automated high availability, and agility as it reacts in real time to changing needs and market scenarios. For testing this pipeline, we train and integrate a YOLOv5 model to detect cotton bolls using the gathered dataset.

\subsection{Lambda Architecture}
Lambda architecture, first proposed in \citep{marz2013big}, is a data processing architecture that addresses the problem of handling both batch and real-time data processing by using a combination of a batch layer and a speed layer. In the context of agriculture, various research studies have implemented Lambda architecture pipelines to process and analyze large amounts of sensor data, such as weather data and crop yields, in order to improve crop forecasting and precision agriculture. Very recent work \citep{Roukh2020,Quafig2022} have demonstrated the feasibility of using a Lambda architecture framework in smart farming. 
\subsection{Cloud Computing}
Previous research on big data pipelines has employed on-premise servers for data processing, while the use of cloud computing can substantially reduce the cost for farmers. Cloud providers, such as Microsoft Azure, offer various data centers to ensure availability and provide better security compared to on-premise servers. We propose the adoption of Microsoft Azure Big Data resources to implement a Lambda architecture pipeline in the agriculture industry. Azure Big Data Pipeline is a cloud-based processing service offered by Microsoft that can be utilized for analyzing, processing, and implementing predictive analysis and machine learning-based decisions for agricultural operations. 

\subsubsection{Azure Data Factory}
Azure Data Factory (ADF) allows for the creation of end-to-end complex ETL data workflows while ensuring security requirements. This environment enables the creation and scheduling of data-driven workspace and the ingestion of data from various data stores. It can integrate additional computing services such as HDInsight, Hadoop, Spark, and Azure Machine Learning. ADF is a serverless service, meaning that billing is based on the duration of data movement and the number of activities executed. The service allows for cloud-scale processing, enabling the addition of nodes to handle data in parallel at scales ranging from terabytes to petabytes. Moreover, one common challenge with cloud applications is the need for secure authentication. ADF addresses this issue by supporting Azure Key Vault, a service that stores security credentials \citep{Rawat2018}. Overall, the use of ADF in our pipeline allows for efficient and secure data processing at scale.

\subsection{Related Work}
Previous studies have utilized traditional pixel-based CV methods, such as OpenCV, to identify cotton bolls based on their white pixel coloring \citep{kadeghe2018real}. Another study has explored the use of YOLOv4 in order to detect cotton blooms and bolls \citep{thesma2022}. 
Moreover, the integration of big data architecture has been suggested in previous research to optimize agricultural operations \citep{Wolfert2017}. Parallel studies have explored the use of Lambda architecture pipelines as a viable approach to process and analyze large amounts of sensor data, such as weather data and crop yield, in order to improve forecasting for specific crops. For instance, Roukh presents a cloud-based solution, named WALLeSMART, aimed at mitigating the big data challenges facing smart farming operations \citep{Roukh2020}. The proposed system employs a server-based Lambda architecture on the data collected from 30 dairy farms and 45 weather stations. Similarly, Quafig integrates a big data pipeline inspired by Lambda architecture for smart farming for the purposes of predicting drought status, crop distributions, and machine breakdowns \citep{Quafig2022}. The study suggests the benefits of flexibility and agility when utilizing a big data architecture. Furthermore, cloud-based solutions have become increasingly popular in agriculture due to their scalability and cost-effectiveness. Another study employs big data in the cloud related to weather (climate) and yield data \citep{Chen2014}.

\subsection{Summary of Contributions and Organization of the Paper}
This paper focuses on the use of Microsoft Azure resources to implement and validate a Lambda architecture High-throughput Phenotyping Big Data pipeline for real-time and batch cotton bloom detection, counting, and visualization. We develop data reduction and processing to transfer useful data and separately train a YOLOv5 object detection model and integrate it into our big data pipeline. The pipeline was thoroughly tested and demonstrated through the analysis of a set of 9000 images. 

Despite existing research work on the use of Lambda architecture and its benefits, there is still a lack of studies that elaborate on the development process and tools to construct this architecture. Moreover, there has been limited research on the application of Lambda architecture utilizing cloud computing resources, as most are server based. \textbf{To the best of our knowledge, there is no previous study that elaborates on the implementation of a big data Lambda architecture pipeline utilizing cloud computing resources, specifically Azure, while integrating advanced machine learning models for plant phenotyping applications}. While big data analytics and cloud computing have become increasingly popular in precision agriculture, the integration of these technologies with Lambda architecture for plant phenotyping (our case, cotton) remains an open research area.  Our approach demonstrates the efficacy of utilizing cloud-based resources for the efficient and accurate analysis of large-scale agricultural datasets.

This paper makes several contributions to the research field, which are listed as follows: 
\begin{enumerate}
    \item Introducing a Lambda architecture pipeline that takes into account batch and real-time processing, providing an efficient and scalable solution for data analysis.
    \item Utilizing cloud computing resources, specifically Microsoft Azure, to improve the performance and reliability of the proposed pipeline.
    \item Demonstrating the actual implementation tools and processes used to build the proposed pipeline, enabling other researchers to replicate and build upon our work.
    \item Integrating a big data pipeline for cotton plant phenotyping, which enables the efficient analysis of large volumes of data and provides new insights into the growth and development of cotton plants.
    \item Contributing a new cotton field dataset to the research community, which is currently limited, enabling other researchers to validate and build upon our findings.
\end{enumerate}

The remainder of the paper is organized as follows: Section 2 describes data retrieval; Section 3 summarizes the implemented Lambda architecture pipeline; Section 4 provides a summary of offline AI-based object detection model training; Section 5 discusses fine-tuning methods to optimize the continuous pipeline run-time and showcases final results, and lastly in Section 6, we discuss areas for future work and concluding remarks.


\section{Dataset}

In this study, we employed our own cotton field dataset to evaluate the proposed pipeline for phenotyping analysis. The cotton field dataset will be further elaborated in subsequent sections, including data collection procedures and data preprocessing steps.

\subsection{Cotton Research Farm}

\begin{figure}[h]
    \centering
    \includegraphics[width=0.35\textwidth]{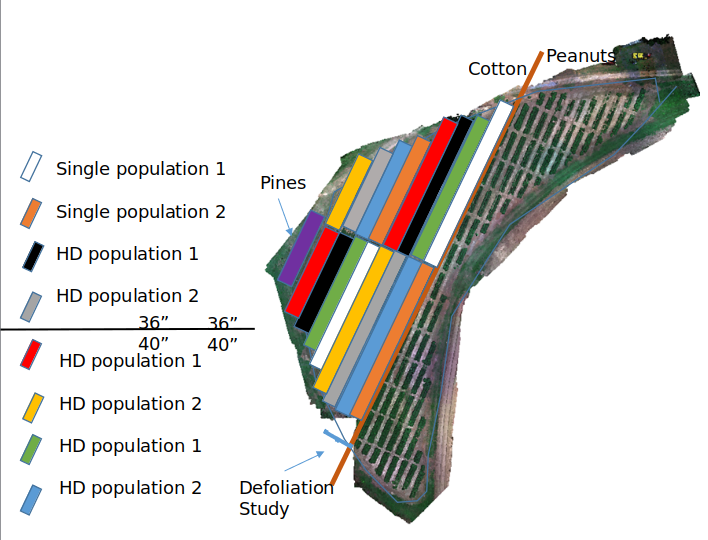}
    \caption{Aerial view of our cotton farm in Tifton, GA displaying 40 rows of cotton plants, treatments of two planting populations (2 and 4 seeds per foot), HD (Hilldrop), and single planted cotton seed. Two-row spacing of 35 inches and 40 inches were also used as treatment. Each treatment was 4 rows wide and 30 feet long. There were three repetitions per treatment.}
    \label{farm}
\end{figure}

The cotton data was collected using a stereo camera that was installed on an autonomous ground vehicle deployed in a research farm at the University of Georgia's Tifton campus in Tifton, GA. Figure \ref{farm} illustrates an aerial view of the farm. The treatments described in Figure \ref{farm} are 4-row wide, but we collected data on the inner 2-rows for post-analysis as discussed later.

\subsection{Cotton Field Data Collection}

In our data collection efforts, we employed a rover developed by West Texas Lee Corp. (Lubbock, Texas). As described in \citep{fue2020center}, this rover is a four-wheel hydrostatic machine with a length of 340 cm, front and rear axles 91 cm from the center, and a ground clearance of 91 cm. It was powered by a Predator 3500 Inverter generator and equipped with the Nvidia Jetson Xavier for remote control and vision and navigation systems. With a top speed of approximately 2 kilometers per hour, the rover was able to efficiently traverse the study area. To power its electronics, the rover utilized two 12-Volt car batteries, as well as a ZED RGB stereo camera.

The ZED stereo camera, with left and right sensors 120 cm apart and mounted 220 cm above the ground facing downward \citep{fue2020center} was chosen for its ability to perform effectively in outdoor environments and provide real-time depth data. It captured 4-5 frames per second and recorded a video stream of each 4-row treatment from June 2021 to October 2021, 2-3 days per week, as a ROS bag file.

\begin{figure}[h]
    \centering
    \includegraphics[width=0.5\textwidth]{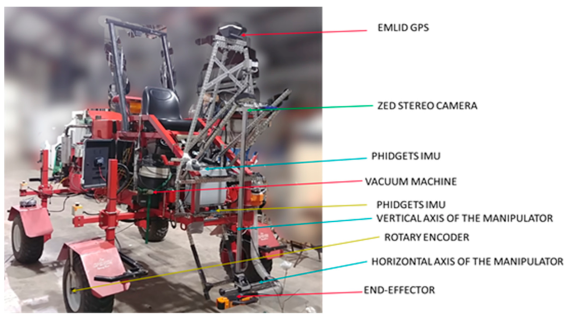}
    \caption{Front view of the rover with the robotic arm, vacuum, and sensors mounted on the rover (see \citep{fue2020center} for details) that was used to collect video streams of cotton plants in Tifton, GA.}
    \label{gv_schem}
\end{figure}

\subsection{Dataset Creation}
\begin{figure}[h]
    \centering
    \includegraphics[width=0.5\textwidth]{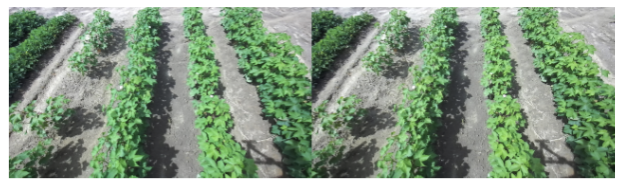}
    \caption{Example of cotton field dataset image after image extraction from original bag files.}
    \label{right-left-lens}
\end{figure}

In this study, a camera equipped with two lenses was utilized to capture images of cotton plants. The camera captured both left and right views of the plants, with a total of 765 image frames extracted from sixteen 4-row treatments on each data collection day between July 14, 2021 and August 6, 2021, where blooms began to appear. These frames were labeled in ascending numerical order to ensure proper correspondence with the video stream and prevent any overlapping. The 765 image frames were subsequently divided into separate sets for the left and right lens views, resulting in a total of 1,530 frames. An example of the image frames captured by the left and right lens can be seen in Figure \ref{right-left-lens}.

\begin{figure}[h]
    \centering
    \includegraphics[width=0.2\textwidth]{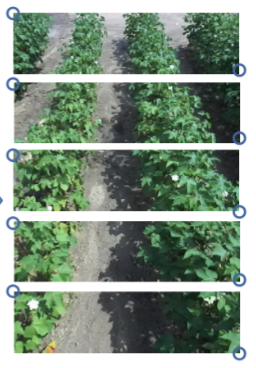}
    \caption{Example of cotton field pipeline input image after preprocessing prior to data ingestation into the pipeline.}
    \label{input-image}
\end{figure}

Previous research has shown that the small proportion of blooms relative to the background in cotton field images can make it difficult for neural network models to accurately detect the blooms \citep{thesma2022}. To address this issue, we pre-processed the images by dividing them into five equal slices. The treatments described in Figure \ref{farm} are 4-row wide, but we collected data on the inner 2-rows for analysis when slicing. An example of the resulting images is shown in Figure \ref{input-image} used as input for the subsequent analysis pipeline. 

We selected a dataset consisting of sliced images from 10 specific days in 2021: July 8, July 14, July 16, July 19, July 23, July 26, July 29, August 4, August 6, and September 9. This resulted in a total of 9,018 images with 3 color channels (RGB) with dimensions of 530 $\times$ 144 for testing batch processing and creating the offline object detection model. The dataset in this study comprised diverse cotton plant data, locations, and treatments, as the video streams were collected from various rows on different days.
\section{Development of Lambda Architecture Pipeline}
\begin{figure}[h]
    \centering
    \includegraphics[width=0.5\textwidth]{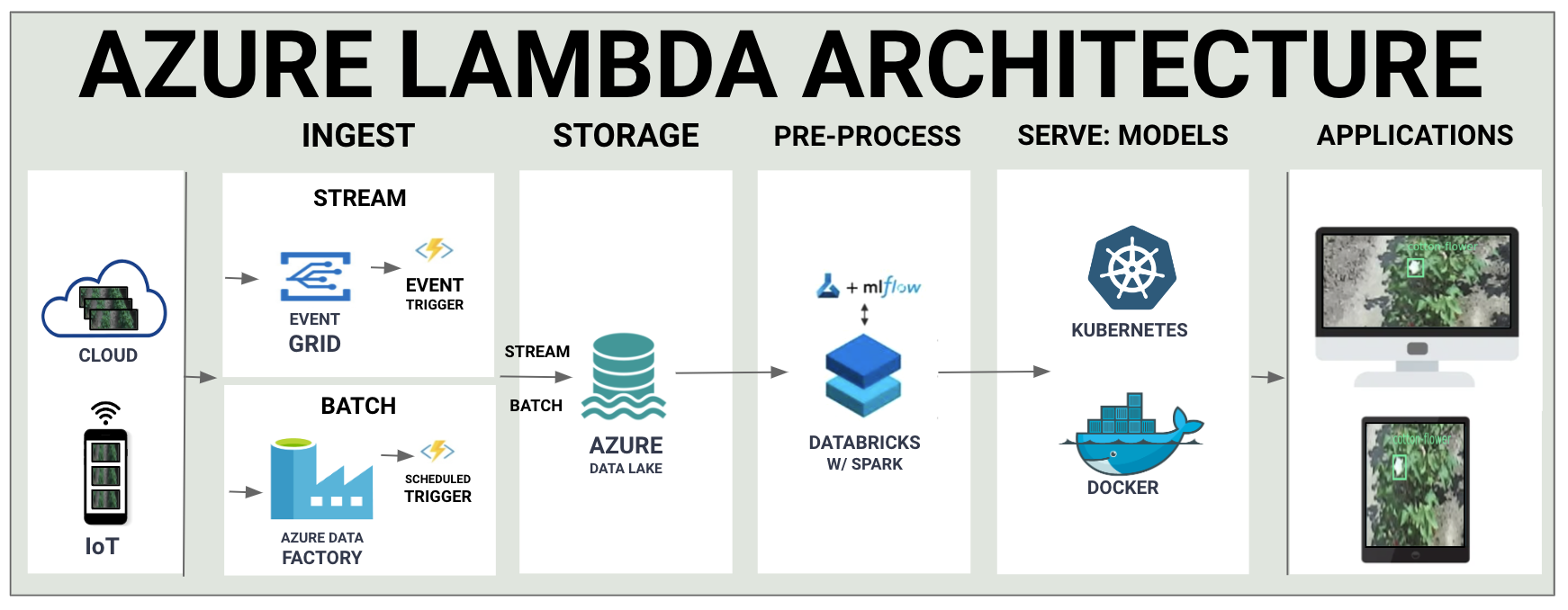}
    \caption{Illustration of the proposed pipeline utilizing Azure resources}
    \label{lambda-architecture}
\end{figure}

In this work, we propose a Lambda architecture to enable real-time analytics through a distributed storage framework, which traditionally is only capable of batch processing. The proposed architecture consists of three main layers: batch, speed, and serving. The batch layer is responsible for processing large amounts of historical data on a schedule, while the speed layer handles real-time streams of data. The serving layer serves the processed data to clients, applications, or users. This approach allows for the efficient handling of both historical and real-time data, enabling a wide range of analytical capabilities. We illustrate the Lambda architecture using Azure resources in Figure \ref{lambda-architecture}. In order to achieve real-time ingestion, we utilize the Azure Data Factory's event-based trigger which sends an event when an image is uploaded to the storage account. This event is handled by Azure's Event Grid for real-time streams. In comparison, batch ingestion is triggered by a scheduled event. Once ingested into the Azure Data Factory, the pipeline connects to Databricks for preprocessing of the image data. The processed data is then forwarded to a deployed AI object detection model, which is running on a Kubernetes cluster, to retrieve the designated bounding box coordinates for the image. Finally, Databricks draws the bounding boxes and outputs the image. The development process will further be elaborated.

To initiate our analysis, we established an Azure Data Factory workspace. The Azure Data Factory portal allows monitoring the pipelines' status in real time. In order to use the Data Factory, we had to create a resource group, a container for holding related resources for our Azure solution. For this work, we opted to ingest binary unstructured data from Azure Blob storage into Azure Data Lake. This allowed us to efficiently process and store large volumes of data for subsequent analysis. 

Azure Blob storage is a highly scalable unstructured data object storage service. To use Blob storage and create an Azure Data Lake, we first had to initialize a storage account. Azure Storage is a Microsoft-managed service that provides cloud storage and provides REST API for CRUD operations. For this project, we configured the storage account to use locally redundant storage (LRS) for data replication, as it is the least expensive option. We also set the blob access tier to `hot' to optimize for frequently accessed and updated data. The storage account's data protection, advanced, and tags settings were left as their default values. Overall, the use of Azure Blob storage and the creation of an Azure Data Lake allowed us to efficiently store and process large volumes of unstructured data for our analysis.

Microsoft Azure Data Lake is a highly scalable data store for unstructured, semi-structured, and structured data \citep{Rawat2018}. It is compatible with Azure services and a variety of additional tools, making it capable of performing data transformation and handling large volumes of data for analytics tasks. To separate the stream and batch processing in our pipeline, we created two separate blob containers labeled batch and stream. Files ingested into the `batch' folder are processed by a scheduled trigger designed for batch processing, while files ingested into the `stream' folder trigger real-time processing. This allows us to efficiently handle both historical and real-time data in our analysis. 

\subsection{Speed Layer} 

The stream layer of the Lambda architecture is designed for real-time analysis of incoming data streams. It is generally not used for training machine learning models, but rather for applying pre-trained models to classify or predict outcomes for the incoming data. This allows to provide real-time insights which are crucial when timely action is required depending on the data. For example, real-time analysis of cotton bloom location and density can enable farmers to take immediate action. Another benefit of the stream layer is its ability to handle high-volume data streams with low latency, which can be a challenge for traditional batch processing systems that may suffer from delays in the availability of insights. 

\subsubsection{Ingestation}

To enable real-time processing in our pipeline, we implemented a file storage trigger in the stream layer. This trigger initiates the pipeline in real time whenever a new image is added to the blob storage. This approach allows us to automate the data processing and analysis pipelines, hence reducing the need for manual intervention. Additionally, the file storage trigger is compatible with other services such as Azure IoT Hub, enabling us to process data ingested from IoT devices for scalability. This approach allows to efficiently and effectively analyze data as it is generated in near real time.

The creation of a real-time trigger in Azure Data Factory also generates an event grid in Azure. Event Grid is a messaging service in Azure that enables the creation of event-driven architectures. It can be used to trigger actions such as running a pipeline. In our case, the event grid listens for events in the input source (blob storage) and, upon detecting a new event, sends a message to the Data Factory service to trigger the execution of the pipeline. This allows for the automation of the pipeline process. For the transfer of data from blob storage into the data lake, we must create a connection between the Data Factory and the Data Lake.  We used a Copy Activity in a Data Factory pipeline to copy data from a Data Lake store to a different store or data sink. 

In our pipeline, we use two separate folders as input sources, each with its own trigger (batch and stream). To facilitate this configuration, we parameterized the input file name to accommodate the separate cases of the stream and batch layers. By adopting the parameterization of the data folder input as dynamic, we were able to alter the folder used as the input source without modifying the pipeline itself. This approach allows us to flexibly configure the input sources for our pipeline without the need for additional maintenance or modification.

\begin{figure}[h]
    \centering
    \includegraphics[width=0.5\textwidth]{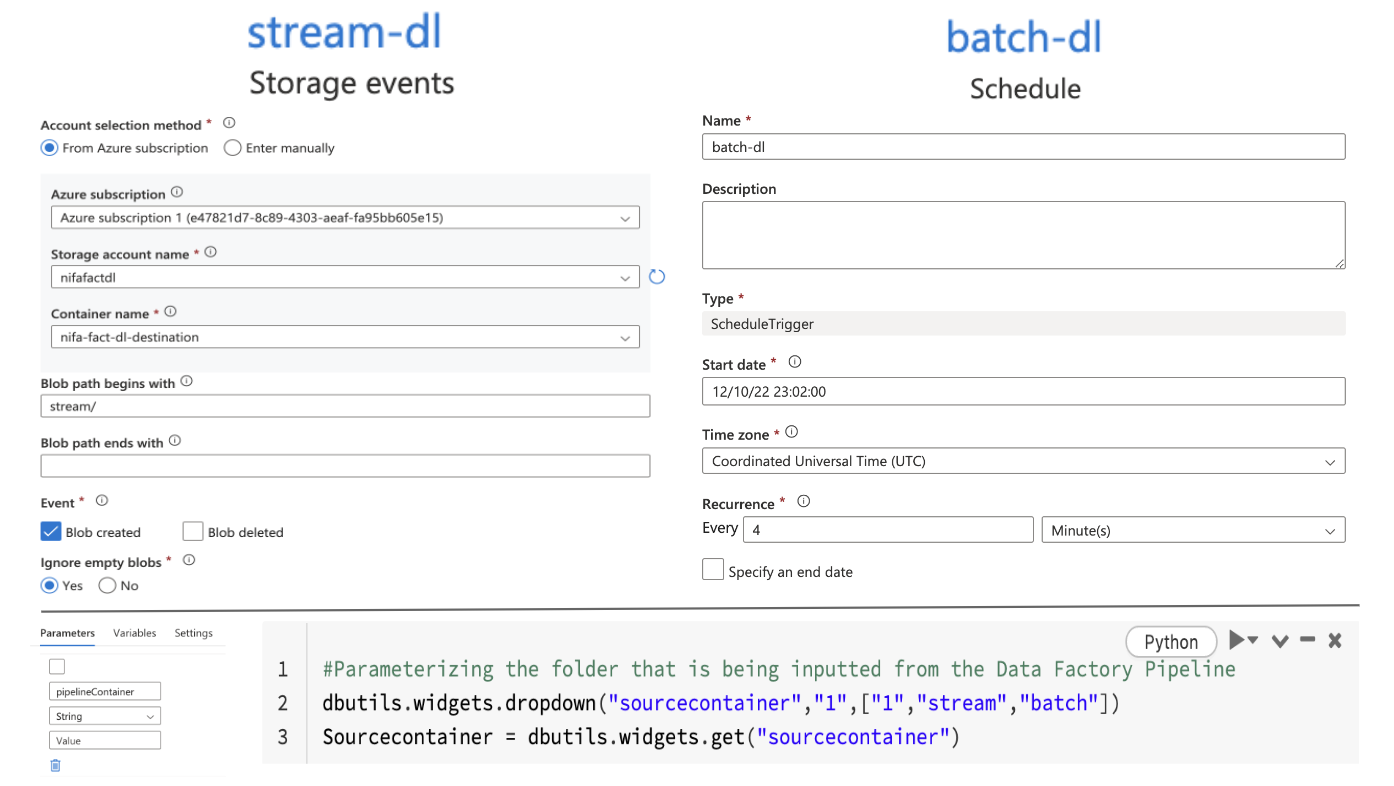}
    \caption{Screenshot of the parameterization process for the stream and batch triggers to automate the pipeline for continuity}
    \label{trigger-parameter}
\end{figure}
\subsection{Batch Layer}

The batch layer of our pipeline serves as the primary repository for the master dataset and allows us to view a batch view of the data prior to computation. The layer plays a crucial role in managing and organizing the dataset, enabling efficient analysis and processing. We can divide this batch data into smaller batches to train machine learning models on large datasets quicker, independently, parallel, and through fewer computational resources. It also helps with the scalability of a machine learning system as the system will be able to handle larger datasets, optimize the training process, and in improving the performance of the resulting model.  

\subsubsection{Ingestation}

For our experiments, we selected a dataset consisting of sliced images from 10 specific days in 2021, which resulted in a total of over 9,000 images. The dataset is stored in Azure Blob Storage, a scalable cloud-based object storage service that is capable of storing and serving large amounts of data. This scalability, compatible with terabytes of data, makes it well-suited for use in data-intensive applications such as ours. To accommodate larger volumes of data, Blob Storage is engineered to scale horizontally by automatically distributing data across multiple storage nodes. This allows it to handle increases in data volume and access requests while eliminating additional manual provisioning or configuration. 

In our pipeline, we integrated a batch trigger in addition to the stream layer trigger. This trigger is of the batch type, allowing us to specify a predetermined schedule for execution. The schedule can be fixed, such as running every day at a specific time, or dynamic through a CRON expression, which is a job scheduler used within Azure. For the purposes of our experimentation, the trigger is calibrated to run every 3 minutes. However, the flexibility of the batch trigger schedule allows for the customization of the frequency of execution to meet our specific data collection and processing needs. For example, the trigger can be executed on a weekly or hourly basis when collecting data on-site. The use of a batch trigger in Azure Data Factory allows us to scalably process large volumes of data. We can ingest data into the batch layer at a rate that meets our specific needs, and schedule the trigger to execute at appropriate intervals to ensure that the data is processed and analyzed in a timely manner. The ability to adjust the schedule of the batch trigger allows us to fine-tune the performance of our pipeline and ensure that it is able to handle the volume and velocity of our data effectively.

\subsubsection{Azure Data Factory Connection}
The batch layer follows the same process as the speed layer for the Azure Data Factory connection. If an image is ingested into the batch folder, the batch trigger sends the parameter of the batch which will be used in the remainder of the pipeline for organizing data. 

\subsection{Pre-process/Analyze}
In the analysis of high-volume data, pre-processing is a vital step. Raw data from devices may contain inconsistencies and noise which can depreciate the quality of results and decision-making insights. These issues are addressed through the cleansing, normalization, and reduction of data. Furthermore, the pre-processing step of images integrates various techniques such as noise reduction, image enhancement, and feature extraction. These methods assist with streamlining decision-making and interpretation. We decide to incorporate image compression into our pipeline as it can significantly reduce the size of images. This downsizes storage size and costs of processing large volumes of data. By integrating image compression methods to eliminate image data redundancy, it is possible to represent an image through fewer bits, resulting in a smaller file size. However, there are trade-offs in terms of image quality and compression ratios, thus it is imperative to select an image compression algorithm that does not deteriorate the quality when compressing. These steps are crucial in the context of big data pipelines, where storage space is often a limiting factor.

\subsubsection{Databricks Connection to Data Lake}

To facilitate pre-processing, we incorporate Databricks, a cloud-based platform that integrates Apache Spark, a powerful open-source data processing engine \citep{zaharia2012}. Apache Spark is optimized to handle substantial amounts of data quickly and efficiently, making it ideal for real-time data processing applications. It boasts the capability for in-memory processing, rendering it significantly more efficient than disk-based systems, especially when working with vast amounts of data, resulting in reduced computation time. Moreover, Apache Spark supports parallel processing, permitting it to divide data into smaller chunks and process them simultaneously to enhance performance even further if needed.

In the realm of pre-processing tasks, a popular alternative to Databricks is the open-source big data processing framework, Hadoop. Hadoop utilizes the MapReduce programming model, which has been shown to be challenging to work with in comparison to the Spark engine utilized by Databricks \citep{gonzalez_2014}. Furthermore, Hadoop requires significant configuration and maintenance efforts to set up and run properly, whereas Databricks offers a user-friendly interface and requires less infrastructure \citep{zaharia2012}. In addition, Databricks provides a range of additional tools and features, such as integration with data storage platforms like Amazon S3 and Azure Blob Storage, as well as the ability for data scientists and analysts to collaborate through notebooks and dashboards \citep{Databricks2021}, making it a more convenient platform for handling big data.

For our experiments, we configured our Databricks cluster to use Databricks Runtime version 11.3 LTS. The worker and driver type is Standard DS3 v2 which contains 14 GB memory and 4 cores. We have the range of workers to be between 2 to 8 and enabled auto scaling, where the cluster configures the appropriate number of workers based on the load.
Once the data is ingested into the Data Lake, we decide to compress the image by storing the image into a jpeg file with a 30\% quality. This pre-processing stage is flexible and scalable where we can also implement other pre-processing and data transformation techniques such as image slicing. Furthermore, we checked the image dimensions to be a valid input for our model. 

The next step is to configure the Databricks linked service connection. The Databricks linked service connection in Azure is a way to connect to a Databricks workspace from Azure. It allows users to easily access and integrate data stored in their Databricks workspace with Azure Data Factory. When configuring the Databricks linked service, we enter the Databricks workspace URL and authentication access token. We first selected the method of having a new job cluster created anytime there was an ingestation trigger. 
\begin{figure}[h]
    \centering
    \includegraphics[width=0.5\textwidth]{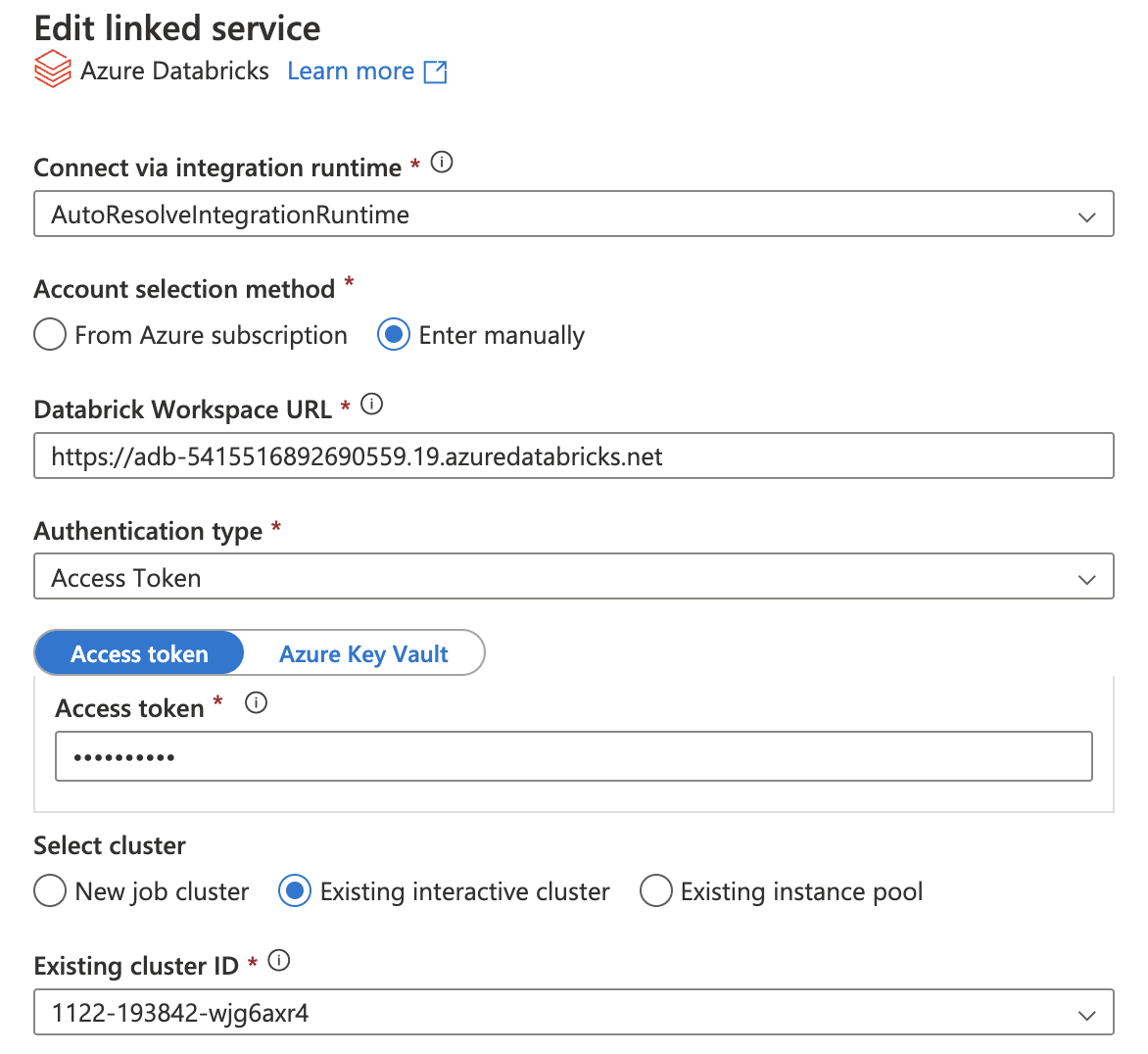}
    \caption{Screenshot of Azure Data factory when setting up the Databricks linked service connection to ADF. The credentials required are as follows: Databricks workspace URL, Authentication type, Access token. Initially, we decided to create a new job cluster; however, based on the results, we shifted to existing interactive cluster; hence, we input the existing cluster ID.}
    \label{databricks-linked-service}
\end{figure}

In order to improve the efficiency of the data processing pipeline, we decided to switch from creating a new job cluster for each ingestion trigger to using existing interactive clusters. This approach reduces the time required for the pipeline to start processing, as the interactive cluster is already active when new data is ingested. This process saves the average 3 minutes of restarting a new cluster for every single trigger. However, when a file is first uploaded, there exists a delay while the inactive interactive cluster is first started. To minimize this delay, we calibrate the interactive cluster to terminate if no activity has been detected for a period of 20 minutes. This configuration can be easily adjusted to meet the needs of different use cases. This results in the first image ingestion taking 3 minutes to begin the cluster, however, subsequent image ingestions demonstrated a significant reduction in connection time to the cluster, with a duration of fewer than 10 seconds.

To enable Databricks to access the Azure Data Factory, we mounted the Data Lake Storage Gen2 (ADLS Gen2) file system to the Databricks workspace. This allows us to use standard file system operations to read and write files in the ADLS Gen2 file system as if it were a local file system. Mounting the ADLS Gen2 file system to Databricks enables us to access data stored in ADLS Gen2 from Databricks notebooks and jobs, and facilitates integration between Databricks and other tools and systems that use ADLS Gen2 as a storage backend. Furthermore, the parameterization of the input folder (batch vs. stream folder) allows the databricks notebook to use this Dynamic input to make changes to the correct data lake folder. 

\subsection{AI Model/APIs}
In this section, we describe the process of deploying the trained AI model.

\subsubsection{Deployment with Kubernetes}

To deploy a trained Object Detection model in the pipeline, we utilized Azure Kubernetes Service (AKS) \citep{azure_kubernetes_service}. Microsoft Azure's AKS simplifies the process of deploying and scaling containerized applications on the cloud platform through its managed Kubernetes service. By leveraging the benefits of the open-source Kubernetes container orchestration platform, AKS creates a consistent and predictable environment for managing these applications. With features like automatic bin packing, load balancing, and secret and configuration management, AKS enhances the management of containerized applications. The service achieves this by creating and managing clusters of virtual machines that run these applications, making the deployment and scaling process easier and more efficient.

Kubernetes is highly scalable, and its platform allows for management of applications across multiple nodes in a cluster, making it a versatile solution for managing containerized applications in the cloud \citep{azure_kubernetes_service}. It provides a consistent and predictable environment for deploying and scaling containerized applications. Secret and configuration management provides secure, encrypted storage for sensitive data such as passwords and API keys, improving application security.  Kubernetes also includes several features that enhance the management of containerized applications, including automatic bin packing, load balancing, and secret and configuration management. Automatic bin packing allows Kubernetes to schedule containers to run on the most appropriate nodes in a cluster, maximizing cost efficiency. Load balancing distributes incoming traffic across multiple replicas of an application to handle high traffic volumes. 

This cluster uses a Standard D3 v2 virtual machine which has 4 cores, 14 GB RAM, and 200GB storage. We retrieve the score Python script from the best AutoML YOLOv5 run, and use it to deploy the model as an AKS web service. In order to assess the performance of our machine learning model, we utilized a Python script. This script contains code for loading the trained model, reading in data, making predictions using the model, and calculating various performance metrics such as accuracy and precision. It also includes provisions for saving the predictions made by the model and the calculated performance metrics to a file or database for further analysis. By running our script on a separate dataset, known as the test dataset, we were able to obtain an unbiased estimate of the model's performance and assess its ability to generalize to new data.

To enhance the capability of our object detection model in handling elevated workloads, it was deployed with autoscaling enabled. This allows for dynamic adjustment of computing resources, such as CPU Processing Nodes and memory, in response to incoming requests. The initial configuration was set to 1 CPU core and 7 GB of memory. To secure the model, an authentication key system was implemented, requiring the provision of a unique key with each request. This key system ensures only authorized access to the model. Subsequent sections of this research will elaborate on the training process of the object detection model and its integration into the workflow.

\subsubsection{Azure Data Factory Connection}
\begin{figure}[h]
    \centering
    \includegraphics[width=0.5\textwidth]{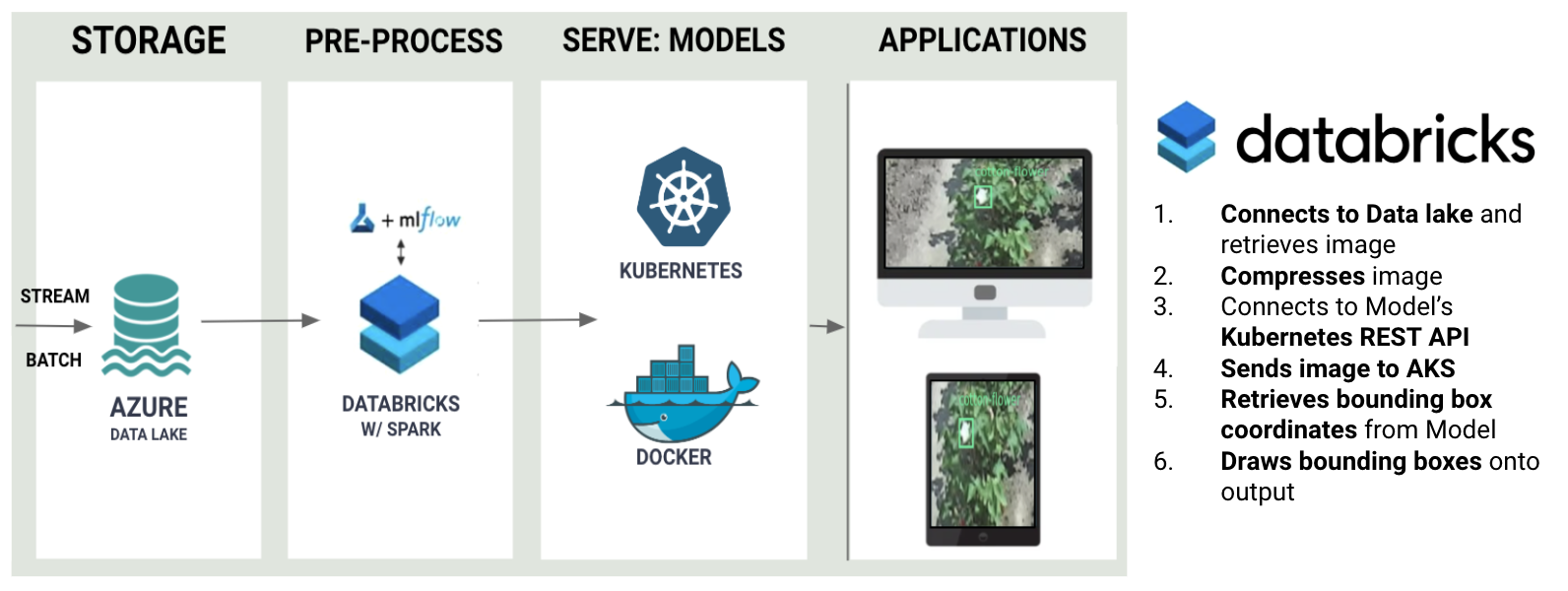}
    \caption{The figure illustrates the tasks within Databricks notebook: compression (pre-processing), connection to AI Model, and creation of output with results (post-processing)}
    \label{databricks-tasks}
\end{figure}

In order to optimize the efficiency of our pipeline, we made the decision to include the AKS connection credentials within the initial Databricks notebook where the data is pre-processed. This approach was chosen as an alternative to utilizing Azure's ADF web service option for REST API connection in the Azure Data Factory, which would have required the creation of another Databricks notebook to draw the bounding boxes from the output of bounding box coordinates. By integrating the AKS connection credentials directly into the primary notebook, we were able to streamline the process while eliminating the need for an additional Databricks compute cluster and cluster connection time. This avoided the added overhead of creating an additional notebook in ADF, which would have slowed down the pipeline. Overall, we have one Databricks notebook which will conduct the pre-processing and post-processing of data. Figure \ref{databricks-tasks} illustrates the tasks of Databricks in the pipeline.

\subsection{Output}
We developed an object detection model that is capable of identifying and counting cotton blooms in images. When the model is run on an input image, it returns the bounding box coordinates of any cotton blooms that it detects. To visualize the results of the model, we retrieve these bounding box coordinates and use them to create visual bounding boxes over the input image. This output image, which shows the detected cotton blooms overlaid on the original image, is then stored in a blob storage account. By using a blob storage REST API, we can easily send this output image to any other device for further processing or analysis. This approach allows us to scale the output of the model to meet needs.

\section{Offline YOLOv5 Model Training}

Object detection is a key task in computer vision, which involves identifying and locating objects of interest in images or video streams. One popular object detection model is YOLO (You Only Look Once), which was first introduced by Redmon in 2015 \citep{redmon2015yolo}. Since then, the YOLO model has undergone several revisions, and one key difference between YOLOv5 and its predecessor, YOLOv4, is the training process. While previous versions of YOLO, including YOLOv4, were trained using the Darknet framework, YOLOv5 utilizes the TensorFlow backend. This allows YOLOv5 to benefit from the advanced optimization and acceleration techniques provided by TensorFlow, which can improve the model's performance and speed. YOLOv5 also introduces several other improvements and new features compared to YOLOv4. These include more efficient network architecture and support for a wider range of input sizes \citep{bochkovskiyyolov4}.

\subsection{Data Labeling}

We utilized AutoML and Azure Machine Learning Studio to train a YOLOv5 model for cotton bloom detection. AutoML automates the process of selecting and training the most suitable machine learning model for a given dataset. It allows users to easily train, evaluate, and deploy machine learning models without the need for extensive programming knowledge or machine learning expertise \citep{wachs2021automl}. To train the YOLOv5 model using AutoML, we first set up a connection between our data lake (which contained the images used for training) and Azure Machine Learning Studio. Azure Machine Learning Studio is a cloud-based platform that provides tools for developing, deploying, and managing machine learning models in Azure \citep{murphy2012machine}. Once the connection was established, we were able to use AutoML and Azure Machine Learning Studio to train and evaluate the YOLOv5 model on our dataset. The platform provided a range of tools and resources for optimizing the model's performance, including the ability to tune hyperparameters, apply data augmentation techniques, and evaluate the model's performance using a variety of metrics which will be further discussed.

\begin{figure}[h]
    \centering
    \includegraphics[width=0.5\textwidth]{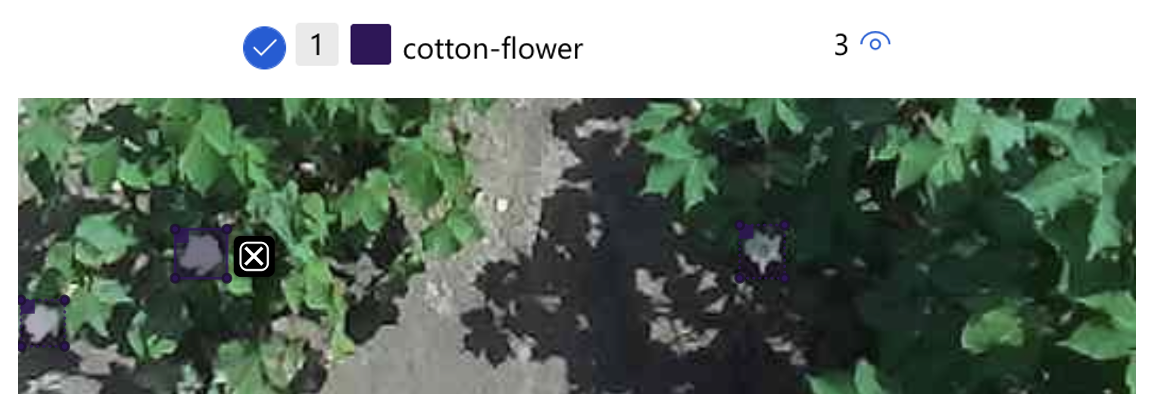}
    \caption{Example of cotton bloom bounding box annotations for one cotton field sliced image.}
    \label{image-labeler}
\end{figure}

After creating the Machine Learning studio workspace, we need to create a Datastore which connects to our Data Lake Storage Container. From the Datastore, we created a Data Asset. Data stores and data assets are resources in Azure Machine Learning studio that allows us to store and access data for machine learning experiments. We created the Data Asset through 1300 images saved in a Data Lake that was compressed prior. In our case, we decided to reduce the quality of the images by compressing prior to training the model. This way, our model would provide better accuracy when implementing the full pipeline which compresses the images prior to being send into the AI Model. 
Using the Azure ML Studio Labeler tool, we annotated 1300 images through bounding boxes that can be used to identify the location and size of the cotton blooms in the image. The AutoML labeler tool is part of the Azure Machine Learning platform. After the annotations were complete, we exported them into AzureML Dataset format. Figure \ref{image-labeler} is a screenshot of an example of annotating one image through Azure's Image Labeler tool. 
\subsection{Model Hyperparameters and Training}

\begin{figure}[h]
    \centering
    \includegraphics[width=0.5\textwidth]{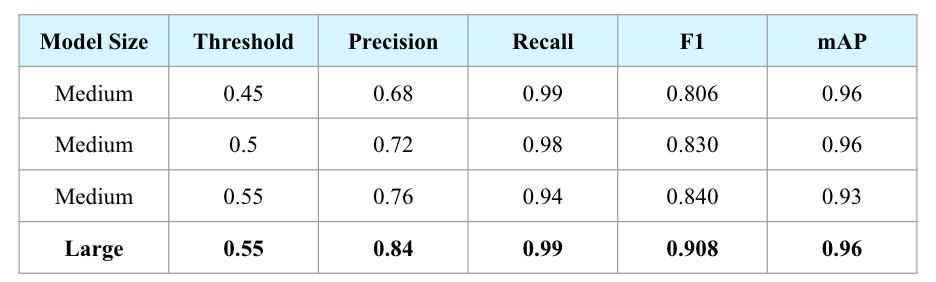}
    \caption{Table displays results from tuning hyperparameters. The F1 score and mAP was the highest when utilizing the large YOLOv5 model with a threshold 0.55. We also tuned the number of epochs, but AutoML would terminate after 30 epochs due to no significant improvement.}
    \label{model-results}
\end{figure}

In this work, the model utilized 80 percent of the dataset for training, and 20 percent for validation purposes. Furthermore, the YOLOv5 model was trained using a learning rate of 0.01, a model size of large which contains 46.5 million training parameters, and a total of 70 epochs. However, the training process was terminated early when the mean average precision (mAP) metrics stopped improving. This resulted in the training process stopping early at 30 epochs in our experiment. The number of epochs used for training is  important, as it determines the number of times that the model sees the training data and can influence the model's performance. Figure \ref{model-results} shows results from our hyperparameter tunings. 

\begin{figure}[h]
    \centering
    \includegraphics[width=0.5\textwidth]{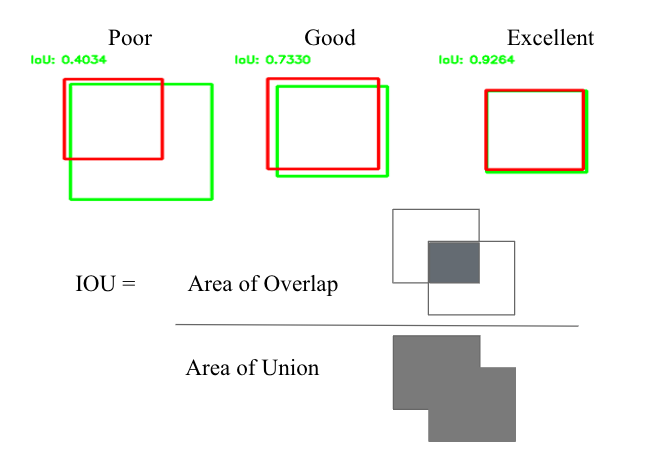}
    \caption{Figure illustrates the definition of IOU which takes into account the area of overlap and the area of union. The higher the area of overlap between the detected bounding box and the ground truth box, the higher the IOU.}
    \label{IOU}
\end{figure}

One key aspect of the training process was the use of the Intersection over Union (IOU) threshold, which is a measure of the overlap between the predicted bounding boxes and the ground truth bounding boxes (see Figure \ref{IOU}). The IOU threshold was set to 0.55 for both precision and recall, which means that a predicted bounding box was considered correct if the overlap with the ground truth bounding box was greater than or equal to 0.55. The use of the IOU threshold is important because it allows the model to be evaluated using a standard metric to compare the performance of different models.
In addition to the IOU threshold, the training process also involved setting the batch size to 10, where the model parameters were updated for each batch of 10 images. This training was performed using a computing cluster with 6 cores, 1 GPU, 56 GB of RAM, and 360 GB of disk space. The overall training process took 1 hour and 10 minutes to complete. \citep{lecun2015deep}.

\section{Finetuning and Results}

\subsection{YOLOV5 Model}

In this work, the trained YOLOv5 AutoML model achieved a mean average precision (mAP) score of 0.96. The mAP score is a metric that is commonly used to evaluate the performance of object detection models. It measures the average precision across all classes of objects in the dataset and takes into account the overall precision and recall of the model. Precision is a measure of the accuracy of the model's predictions and defined as the number of correct predictions divided by the total number of predictions. In comparison, recall calculates the model's ability to capture all relevant instances in its predictions. It can be determined by dividing the number of correct predictions by the total number of instances in the actual data \citep{lecun2015deep}. 

In this case, the YOLOv5 model had a precision value of 0.84 and a recall score of 0.99 when using an IOU validation threshold of 0.55. The F1 score, which is a measure of the harmonic mean of precision and recall, was also calculated and found to be 0.904. The importance of precision, recall, and the F1 score lie in their ability to provide a comprehensive evaluation of the model's performance. High precision is essential for ensuring that the model does not produce false positives. A high recall is essential for ensuring that the model does not produce false negatives. The F1 score, which takes into account both precision and recall, provides a balanced evaluation of the model's performance \citep{lecun2015deep}. Below displays the formulas mentioned above which consider the True Positive (\textit{TP}), False Positive (\textit{FP}), and False Negative (\textit{FN}): 
\begin{eqnarray}
 \mathrm{precision}&=\frac{TP}{TP+FP}\\
 \mathrm{recall}&=\frac{TP}{TP+FN}\\
 \mathrm{F1\ Score}&=\frac{2\times \rm{precision\times recall}}{\rm{precision+recall}}
\end{eqnarray}

\begin{figure}[h]
    \centering
    \includegraphics[width=0.5\textwidth]{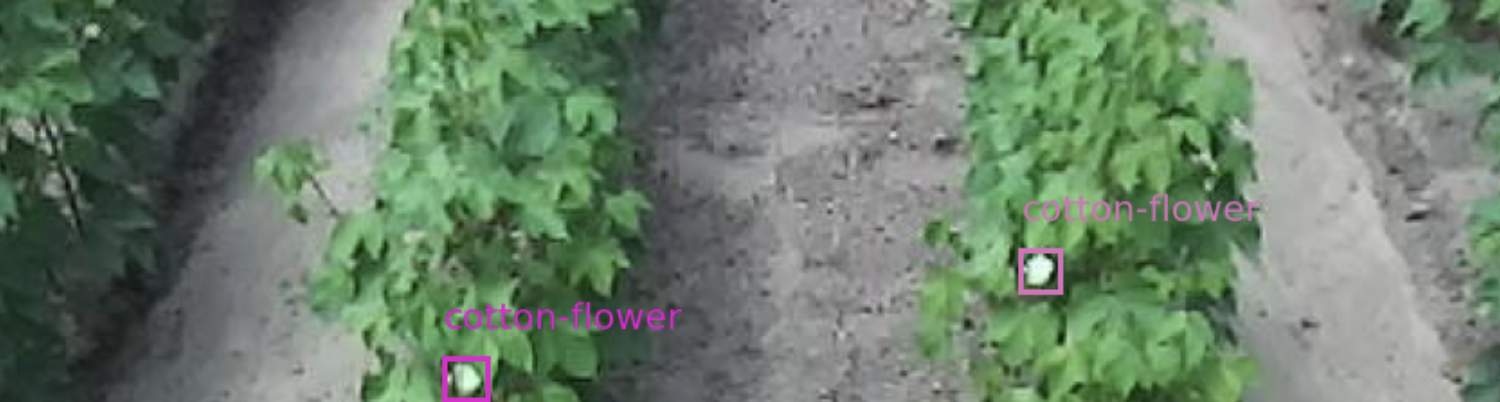}
    \caption{Example of pipeline output after post-processing and adding bounding boxes for cotton bloom detection visualization.}
    \label{post-processing}
\end{figure}

The model itself returns back the bounding box coordinates. When integrating the model into the pipeline, we conduct post-processing to draw and visualize the bounding boxes on top of the input image. Figure \ref{post-processing} displays the output of the cotton bloom detected image from the AI model.

\subsection{Azure Data Factory}

In order to connect the trained AI model into the rest of the Azure Data Factory Pipeline, we first created a Standard Kubernetes cluster. We then deployed the model into Kubernetes which provides a REST API to interact with.

\subsubsection{REST API vs. Blob Storage Ingestation}

Previously, we ingested data from Azure Blob Storage into Azure Data Lake to demonstrate the feasibility of ingesting data from external IoT devices. To assess the performance of the ingestion process, we conducted an experiment using image data and the REST API connection provided by the Data Lake. Initially, we utilized the popular API development and testing tool, Postman, to conduct a synchronous request and observed a substantial improvement in ingestion time. It is commonly used for testing and debugging API applications, and can be used to make both synchronous and asynchronous requests \citep{fielding2000architectural}. The implementation of this reduced the stream ingestation time from 12 seconds to 150 ms. This not only highlights the applicability of the REST API connection but also its efficiency in speeding up the ingestion process.

While Postman is a useful tool for testing and debugging APIs, it is not the only option for making HTTP requests to devices. To scale up for batch processing, we adopted asynchronous Python code for HTTP connection. The original ingestation time from blob storage to the data lake took around 2 minutes for 9,000 images (157.6 MB). With the optimization of the REST API and asynchronous Python code, the batch ingestion process was completed in just 8.62 seconds, a marked improvement from the previous ingestion time.

For future purposes, one can use the REST API and HTTP connection with other devices and systems (mobile devices or IoT devices). The pipeline is compatible with the integration of machine learning models into a wide range of applications and systems.

\subsubsection{Kubernetes}

We also optimized our Kubernetes configurations by increasing the number of nodes and node pools. When testing on a smaller batch amount of 100 images, integrating 5 nodes rather than 3 nodes in the Kubernetes cluster decreases runtime from 32 minutes to 28 minutes. Increasing the number of node pools from 1 pool to 2 pools decreased runtime from 28 minutes to 22 minutes.

\subsubsection{Asyncronous vs. Synchronous Processing}
Asynchronous programming allows the execution of multiple tasks to run concurrently without waiting for the completion of prior tasks. The asyncio library, a built-in library in Python, provides the infrastructure for writing asynchronous code with the help of the async and await keywords \citep{asyncio_library}. Additionally, the aiohttp library enables asynchronous support for HTTP requests, allowing for concurrent processing of multiple requests without waiting for responses {\citep{aiohttp_library}. 

The aiofiles library, on the other hand, offers asynchronous support for file operations such as reading and writing to files. This can be useful in programs that need to perform numerous file operations simultaneously, such as our program that handles a significant amount of images \cite{aiofiles_library}. Our pipeline runtime for processing 9,000 batch images was found to take approximately 3 hours and 50 minutes with synchronous code. After optimizing the pipeline with asynchronous code, the execution time was reduced to 34 minutes, which represents a substantial improvement. This demonstrates the potential benefits of implementing asynchronous processing in our pipeline 
\subsection{Cost}

Although Azure is not an open-source environment, the pay-as-you-go service makes sure to charge resources that are effectively used. With Microsoft Azure, we can spin a 100-node Apache Spark Cluster in less than ten minutes and pay only for the time the job runs on the specific cluster \citep{Rawat2018}.

We used a computer cluster that had the GPU infrastructure for the YOLOv5 training. This costs \$1.14 per hour. The total time spent training was 1 hour and 6 minutes. The total cost is as follows: using the virtual machines led to a cost of about \$3.56, storage cost \$2.18, container costs were \$1.85, utilizing a virtual network was \$1.33, Azure Databricks connection was \$0.30, and Azure Data factory led to an additional cost of \$0.30. Furthermore, the Kubernetes cluster deployment was the most costly item as ranges roughly about \$70 monthly. 

\section{Future Directions}

The pipeline can be further optimized by updating computing clusters with higher computing power and incorporating GPU processing to reduce the total processing time. Moreover, the pipeline currently takes approximately three minutes to reactivate the terminated Databricks interactive cluster, which could be improved through the use of pools. 

A bottleneck encountered during the data processing was the connection to the Internet to send images to the Kubernetes cluster through REST API. To address this issue, we can utilize Databricks MLFlow by downloading the model within the Databricks environment itself rather than having to create a separate Internet connection. We refer back to Figure \ref{databricks-tasks} to gain a better understanding of the bottleneck at step 3 to where the cluster must create an Internet connection to the REST API URL. If we wanted to scale up with more nodes, the price of Kubernetes would increase to even up to \$1,000 monthly. This further suggests the benefit of utilizing Databricks MLFlow and downloading the model itself rather than using Kubernetes' REST API for AI Model connection. Another bottleneck encountered is the limitations of OpenCV when drawing bounding boxes. Despite our efforts to optimize results through asynchronous Python code, OpenCV does not have the capability for asynchronous processing. As a result, it is incapable of performing the task of producing bounding boxes concurrently for images. This is because OpenCV relies heavily on the CPU, which is fully operated without waiting for any external input. This results in a linear process when drawing bounding boxes, despite the rest of the code being optimized for concurrent execution. To overcome this issue, we can incorporate PySpark, a Python library for distributed data processing using Apache Spark. PySpark allows us to leverage the power of Spark, which is a distributed computing platform that enables fast and flexible data processing. This is compatible with our pipeline because our Databricks runtime version 11.3 LTS includes Apache Spark 3.3.0, and Scala 2.12. With the use of PySpark, we can employ the parallel computing power of our Databricks cluster and enhance the speed and efficiency of our data processing operations.

Overall, these optimization strategies can be used to scale up the pipeline and decrease the total processing time, making it more efficient and effective for handling much larger datasets.

\section{Conclusion}
This study has presented a new big data pipeline for cotton bloom detection using a Lambda architecture and Microsoft Azure's cloud computing resources. The pipeline fulfills data preprocessing, object detection using a YOLOv5 neural network trained through Azure AutoML, and visualization of object detection bounding boxes. The results of the study demonstrate the high performance of the neural network with a Mean Average Precision (mAP) score of 0.96 and an optimized runtime of 34 minutes when evaluated on over 9,000 images. This work showcases the scalability of the presented pipeline as a solution for deep learning-based object detection and emphasizes on the potential of employing cloud computing resources for big data processing in precision agriculture. This study advances the field by expanding and demonstrating the big data pipeline implementation of a new method for cotton bloom detection from images collected on a cotton farm. The results obtained in this study suggest a scalable Lambda architecture that can be implemented for big data processing using Azure resources.

\section*{Acknowledgement}

The authors would like to thank Canicius Mwitta for his assistance in setting up the experiments and data collection.

\bibliographystyle{cas-model2-names}

\bibliography{cas-refs}
\end{document}